\documentclass[sigconf,nonacm]{acmart}

\AtBeginDocument{%
  }

\begin{document}

\title[Semantic and Spatio-Temporal Co-evolution for Edge Classification]{An Analysis of Semantic and Spatio-Temporal Co-evolution for Edge Classification on Dynamic Text-Attributed Graphs}

\author{Tyler Bonnet}
\email{t.bonnet24@imperial.ac.uk}
\affiliation{%
  \institution{Imperial College London}
  \city{London}
  \country{United Kingdom}
}

\author{Marek Rei}
\email{marek.rei@imperial.ac.uk}
\affiliation{%
  \institution{Imperial College London}
  \city{London}
  \country{United Kingdom}
}

\renewcommand{\shortauthors}{Bonnet and Rei}

\begin{abstract}
Dynamic edge classification on text-attributed Web graphs, such as review networks, community forums, and news-derived knowledge graphs, requires high quality, adaptive semantic representations in the context of network evolution. Existing dynamic graph architectures typically initialize text feature vectors with pretrained language models and hold them static during training to preserve their semantic structure, forgoing the potential benefits of allowing them to co-evolve with network topology. In this work, we investigate the effects of enabling such co-evolution with a unified text-graph alignment framework that integrates a lightweight Transformer adapter to update token-level text representations via spatio-temporal gradients from a contrastive pretraining objective. To prevent detrimental semantic distortion, a language model maintenance objective anchors the evolving text embeddings to their original semantic structure. Evaluations on the Dynamic Text-Attributed Graph Benchmark demonstrate that this anchored text-graph co-evolution consistently improves upon the edge classification performance of frozen-text baselines across diverse Web-related domains. Representational analyses reveal that the framework reduces anisotropy, improving the linear separability of text embeddings, and effectively encodes structural and temporal information into the semantic space. A computational efficiency analysis confirms that this beneficial semantic and spatio-temporal co-evolution is achieved without incurring the prohibitive computational overhead typical of end-to-end joint
training.
\end{abstract}


\maketitle

\section{Introduction}
Evolving networks of the Web, where interactions carry rich text such as product reviews and community Q\&A, can be represented as dynamic text-attributed graphs. Edge classification on these graphs drives Web applications such as predicting review ratings and filtering useful community forum responses. These tasks require capturing dependencies between textual content and spatio-temporal dynamics. By studying how to represent text within an evolving network, our work addresses representation learning on temporal, attributed Web-related graphs and falls within the scope of the \textit{Graph Algorithms and Modeling for the Web} track. Continuous-Time Dynamic Graph Networks (CTDGNs) have become standard for such applications, processing interactions as streaming sequences to capture dependencies \cite{kumar2019jodie, trivedi2019dyrep, xu2020tgat, rossi2020temporal}. Historically, the effective use of text attributes in dynamic graph modeling was not a primary research focus, in part due to the limited availability of datasets that contained such attributes. As a result, subsequent CTDGNs \cite{wang2021cawn, wang2021tcl, cong2023graphmixer, yu2023dygformer} were designed with the assumption that text processing was secondary or handled externally. Such models remain standard baselines in dynamic text-attributed graph modeling \cite{zhang2024dtgb}.

To apply these architectures to text-rich domains, the prevailing approach is to initialize text attributes with a pretrained language model (PLM) and use them as static features during training. While evaluations in DTGB confirm that the frozen text embeddings offer a performance improvement over random vectors, this design remains fundamentally limited. The predictive value of the text in an interaction often depends heavily on the evolving historical topology between the participating nodes. By freezing the text representations, the model is forced to map generic semantic spaces to dynamic structural patterns. Consequently, the text attributes remain entirely decoupled from temporal interaction patterns, forcing the graph encoder to bridge a substantial representational gap alone and ultimately restricting dynamic edge classification performance.

Motivated by these limitations, we investigate the effects of enabling the co-evolution of text representations and network topology in dynamic text-attributed Web graphs with a unified text-graph alignment (UTGA) framework\footnote{Code will be made available upon publication.}. Our approach integrates a dynamic graph encoder with a trainable language model adapter that receives gradient updates from a contrastive pretraining objective, encoding spatio-temporal information into the text embeddings. To prevent detrimental semantic distortion, we use a language model maintenance loss that anchors the outputs of the adapter to the original PLM embeddings.

Evaluation on DTGB demonstrates that UTGA consistently outperforms baselines relying on static text embeddings for dynamic edge classification across a range of label budgets. An analysis of the learned textual latent space shows that UTGA reduces anisotropy, substantially improving the linear separability of the text embeddings, and effectively encodes structural and temporal properties into the semantic space. Crucially, this co-evolution avoids the prohibitive computational overhead typical of end-to-end training, operating within the same memory footprint order of magnitude as frozen-text baselines.

\section{Related Work}

This section reviews text attribute processing in standard CTDGNs and semantic alignment on text-attributed graphs.

\subsection{Text Attribute Processing in CTDGNs}

The prevailing approach to processing text attributes in CTDGNs was established by JODIE~\cite{kumar2019jodie}, which treats them as fixed inputs rather than learnable representations in its state update equation as it compresses historical interactions into latent states. DyRep~\cite{trivedi2019dyrep} adopts this strategy, incorporating fixed representations of text attributes to update latent node states, which in turn parameterize an intensity function to modulate event occurrence rates. This treatment of text was formalized by TGN~\cite{rossi2020temporal}, which unified memory-based methods into a generic framework where text representations continued to serve as static components of the message vector passed to the recurrent module. In these architectures, learning is confined to the spatio-temporal dynamics of the memory update, leaving semantic representations immutable throughout training.

The development of more expressive architectures improved spatio-temporal modeling but retained this passive treatment of text attributes. TGAT \cite{xu2020tgat} aggregates interactions with temporal self-attention, while CAWN \cite{wang2021cawn} extracts network motifs by sampling causal anonymous walks and encoding them with an RNN. Despite these innovations, both architectures process text representations as fixed inputs. In Transformer-based models such as DyGFormer \cite{yu2023dygformer} and TCL \cite{wang2021tcl}, interaction histories are processed as sequences where text embeddings are combined with other feature vectors, such as time delta encodings. Although these models introduce mechanisms such as patching to capture long-range dependencies or contrastive objectives to improve robustness, they do not update the text representations. This paradigm is further cemented by DTGB \cite{zhang2024dtgb}, which, despite shifting attention towards the textual modality of dynamic graphs, standardizes an evaluation protocol that relies on frozen text embeddings.

\subsection{Semantic Alignment on Text-Attributed Graphs}

Semantic alignment strategies for text-attributed graphs can be organized into two broad categories: (1) those that rely on joint-optimization driven by downstream task labels; and (2) task-agnostic self-supervised pretraining frameworks. While there have been recent advancements in the former category in both the static \citep{zhao2023glem, zhu2025promptgfm} and dynamic \citep{zhang2025cross, roy2025lkd4dytag, xu2026moment, wang2026dygrasp} domains, our work is positioned in the latter. Therefore, our review focuses on methods with a decoupled self-supervised pretraining phase.

Early approaches in the static graph domain reduced text to bag-of-words \cite{sen2008collective} or TF-IDF \cite{yang2015tadw} vectors, but with the advent of PLMs came a shift towards deep semantic alignment. GIANT \cite{chien2022giant} aligns semantic and topological spaces by instructing the text encoder to solve a self-supervised neighborhood prediction task, effectively distilling neighborhood structure into the semantic embeddings. To capture more complex topological semantics, Patton \cite{jin2023patton} pretrains language models with network-contextualized masked language modeling and masked node prediction, grounding token representations in the text of adjacent nodes. Because jointly training large language models and graph structures is computationally costly, other lines of research prioritized efficiency. For instance, GraphAdapter \cite{huang2024graphadapter} uses a GNN as a lightweight, trainable adapter to fuse structural representations directly into the hidden states of frozen PLMs, pretraining the framework via task-agnostic next-token prediction. More recently, the field has expanded its focus from single-domain alignment to cross-domain transferability. GraphCLIP \cite{zhu2025graphclip} addresses this by applying a self-supervised contrastive objective that aligns tunable graph encoders with LLM-generated text summaries in a shared semantic subspace. Overall, these approaches demonstrate that actively aligning semantic and structural spaces through end-to-end pretraining or parameter-efficient adaptation is essential to overcome the limitations of frozen-text architectures on static graphs.

In the dynamic graph domain, however, deep semantic alignment remains relatively underexplored. While self-supervised pretraining frameworks spanning generative \cite{liu2025dygmae, gao2025dvgmae} and contrastive \cite{xu2023cldg, zhu2024idol, bonnet2026dygnrole} approaches have advanced generalizable representation learning for dynamic graphs, they lack mechanisms for semantic alignment and retain the inherent limitations of frozen text embeddings.

\section{Methodology}

\begin{figure*}[ht]
    \centering
    \includegraphics[width=\textwidth]{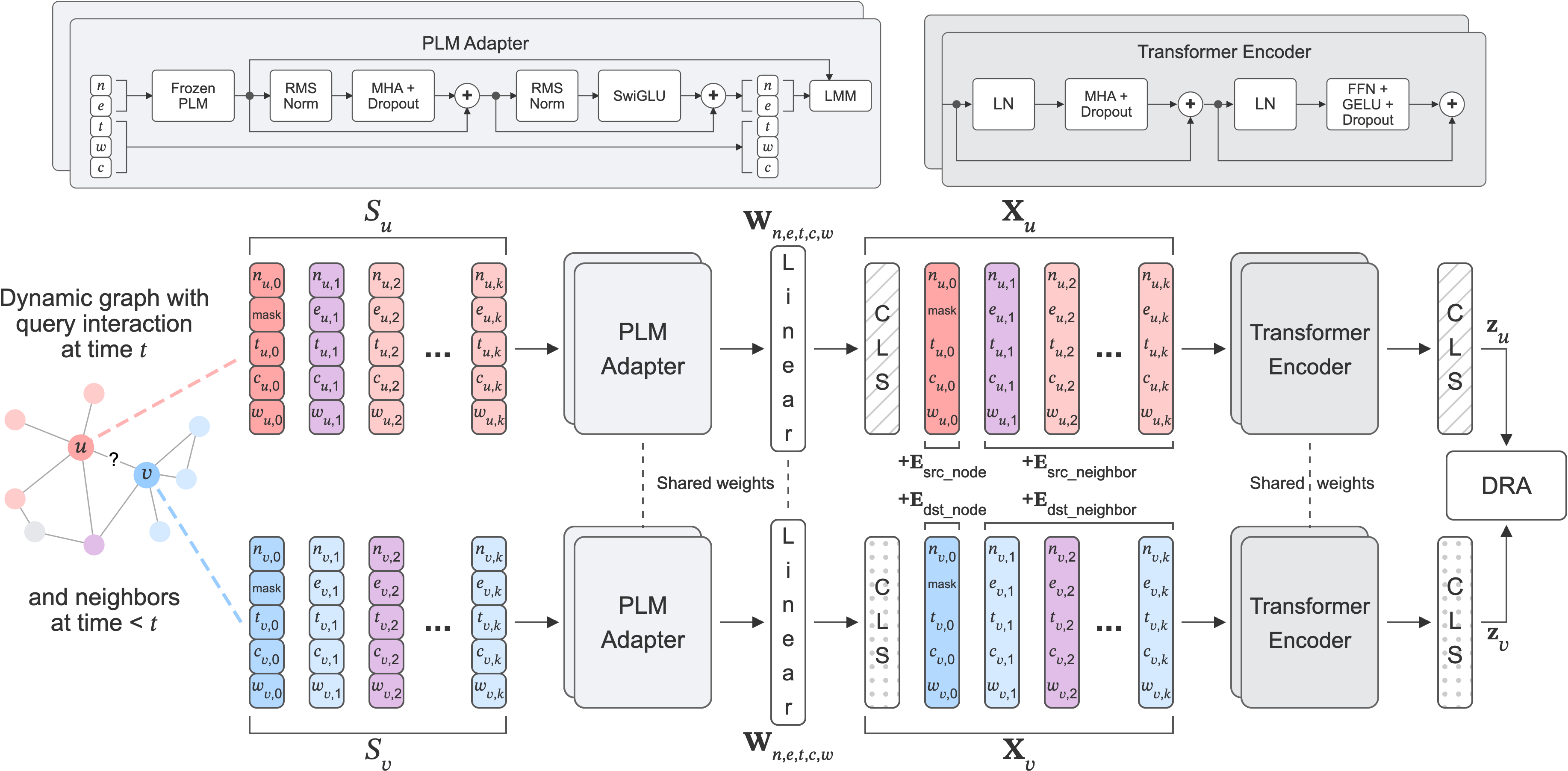}
    \caption{\textbf{Overview of the Unified Text-Graph Alignment Framework.} \textbf{(Left)} For a query interaction between nodes $u$ and $v$ at time $t$, historical one-hop neighbors are retrieved. $u$ and $v$ combined with their respective neighbors form node sequences $S_u$ and $S_v$. For each event in these sequences, its raw node text attribute $n$, raw edge text attribute $e$, interaction time delta $t$, cross-sequence neighbor frequency $c$, and within-sequence neighbor frequency $w$ are extracted, and the edge text attribute of the query nodes is masked. \textbf{(Center)} $n$ and $e$ are processed by a frozen PLM into raw token-level text embeddings, which are subsequently transformed by the PLM adapter. For each node and edge text sequence, the final hidden state corresponding to the EOS token is extracted to serve as the feature vector representing the full text. To prevent semantic collapse, these adapted text feature vectors are anchored to their corresponding raw text embeddings via the language model maintenance (LMM) objective. The text feature vectors and the bypassed temporal and structural feature vectors ($t, c, w$) are independently projected via separate linear layers $\mathbf{W}_{n,e,t,c,w}$ and concatenated. Role-semantic positional encodings (RSPE) are added to the projected vectors to distinguish source query, destination query, source neighbors, and destination neighbors. Distinct CLS tokens are prepended to form the final augmented input matrices $\mathbf{X}_u$ and $\mathbf{X}_v$. \textbf{(Right)} The sequences $\mathbf{X}_u$ and $\mathbf{X}_v$ are processed independently by the Transformer encoder, using shared weights. The CLS tokens are extracted to serve as the final source and destination representations $\mathbf{z}_u$ and $\mathbf{z}_v$, which are optimized via the directional role alignment (DRA) objective.}
    \Description{A block diagram of the Unified Text-Graph Alignment framework.}
    \label{fig:utg_architecture}
\end{figure*}

This section establishes preliminaries and details the UTGA framework (Figure~\ref{fig:utg_architecture}), including the base framework, text adaptation module, and language model maintenance objective.

\subsection{Preliminaries}

A dynamic text-attributed graph is defined as $\mathcal{G} = (\mathcal{V}, \mathcal{E}, \mathcal{T}, \mathcal{D}, \mathcal{R})$, where $\mathcal{V}$ is the node set, $\mathcal{E} \subseteq \mathcal{V} \times \mathcal{V} \times \mathcal{T}$ contains timestamped edges, $\mathcal{T}$ is the set of timestamps at which edges occur, $\mathcal{D}$ is the set of node text attributes, and $\mathcal{R}$ is the set of edge text attributes. An edge $(u, v, t) \in \mathcal{E}$, with its associated node and edge text attributes, connects a source node $u$ to a destination node $v$ at time $t$, and has a label $y_{uv}^t$ representing the edge class.

At $t$, with a set of edges $\mathcal{E}_{\leq t} = \{(u, v, t', y_{uv}^{t'}) \in \mathcal{E} \mid t' \leq t \}$, the evaluation follows the dynamic edge classification protocol of DTGB \cite{zhang2024dtgb}, where the task is defined as predicting the label of an edge at $t+1$ given historical information, with the objective of learning a function $f: (\mathcal{V}, \mathcal{E}_{\leq t}, \mathcal{D}, \mathcal{R}) \rightarrow y_{uv}^{t+1}$, which predicts edge classes between node pairs at future timestamps.

\subsection{Base Framework}
\label{sec:base_framework}

While our unified text-graph alignment framework is inherently encoder-agnostic, we adopt the role-aware sequence modeling framework of DyGnROLE \cite{bonnet2026dygnrole} as our experimental vehicle. This framework belongs to the broader family of Transformer-based CTDGNs that process the historical interactions of query nodes as a sequence \cite{wang2021tcl, yu2023dygformer}. Given a query interaction $(u,v,t)$, the encoder constructs temporal neighborhood sequences for the source and destination nodes. Each node position in the sequence is represented by a concatenation of five feature channels: node text features, edge text features, time encodings, within-sequence frequency embeddings, and cross-sequence frequency embeddings. These representations, along with prepended role-specific CLS tokens, are processed by a Transformer encoder to extract the final source and destination node representations, $\mathbf{z}_u$ and $\mathbf{z}_v$.

To optimize these representations and structure the latent space around the directional interaction process, the framework uses the directional role alignment (DRA) objective. DRA formulates pretraining as an InfoNCE \cite{oord2018representation} retrieval task formulated as
\begin{equation}
\mathcal{L}_{\mathrm{DRA}}
= -\frac{1}{N} \sum_{i=1}^{N} \log \frac{\exp\!\left(\frac{\mathbf{z}_u^{(i)} \cdot \mathbf{z}_v^{(i)}}{\tau}\right)}{\sum_{j=1}^{N} M_{ij} \exp\!\left(\frac{\mathbf{z}_u^{(i)} \cdot \mathbf{z}_v^{(j)}}{\tau}\right)},
\end{equation}
where $N$ is the batch size, $\tau$ is the temperature parameter, and $\mathbf{z}_u^{(i)}$ and $\mathbf{z}_v^{(i)}$ represent the $\ell_2$-normalized embeddings of the source and destination nodes. To prevent penalizing established structural relationships, DRA applies a binary historical mask $M_{ij} \in \{0,1\}$ that removes previously observed interactions from the denominator. 

By default, the DyGnROLE framework relies on static text features generated by a frozen language model. In the following sections, we detail how our approach replaces these static inputs with a fully differentiable text adaptation module, using the gradients from $\mathcal{L}_{\mathrm{DRA}}$ to drive the semantic and topological co-evolution.

\subsection{PLM Adapter}
\label{sec:plm_adapter}
Our framework replaces static features with a fully differentiable PLM adapter, which is a lightweight Transformer \cite{vaswani2017attention} operating on token-level representations produced by a frozen PLM. In our implementation, we use Qwen3-Embedding-0.6B \cite{zhang2025qwen3}, although the proposed adapter architecture is agnostic to the choice of PLM and can be applied to any model capable of producing contextualized token representations. Node and edge attributes are first tokenized and passed through the frozen PLM, from which the full sequence of final-layer hidden states are extracted. Learnable absolute positional embeddings are added to these contextualized token representations, which are then processed by the adapter, providing a trainable semantic transformation between the PLM latent space and the graph encoder.

The adapter consists of two Transformer layers, each containing bidirectional multi-head self-attention followed by a position-wise SwiGLU feed-forward network \cite{shazeer2020swiglu}. Both sub-layers use residual connections and pre-layer normalization using RMSNorm \cite{zhang2019rmsnorm}. For an input sequence of token representations $H$, the adapter computes
\begin{equation}
H' = H + \text{Dropout}(\text{MHA}(\text{RMSNorm}(H)))
\end{equation}
\begin{equation}
H'' = H' + \text{SwiGLU}(\text{RMSNorm}(H')),
\end{equation}

where $\text{MHA}$ denotes multi-head self-attention and $\text{RMSNorm}$ denotes Root Mean Square Normalization. The feed-forward transformation is defined as
\begin{equation}
\text{SwiGLU}(H') =
(\text{SiLU}(H'W_{\text{gate}}) \odot H'W_{\text{val}})W_{\text{out}},
\end{equation}

where $\text{SiLU}$ is the Sigmoid Linear Unit activation function, $W_{\text{gate}}$, $W_{\text{val}}$, and $W_{\text{out}}$ are learnable projection matrices, and $\odot$ denotes element-wise multiplication. Following the adapter transformation, the final hidden state corresponding to the EOS token is extracted as the resulting text representation $\mathbf{z}_{text} = H''_{\text{EOS}}$.

This representation is then passed through the feature channels of the graph encoder, enabling structural gradients from the DRA objective to propagate through the adapter and adapt the semantic space according to temporal interaction patterns.

\subsection{LMM Objective}
\label{sec:lmm_objective}

To prevent detrimental semantic distortion during DRA optimization, we introduce a language model maintenance (LMM) loss. Exposing a text encoder purely to structural gradients risks an unconstrained adaptation that prioritizes network topology at the expense of broader semantic relationships. While standard mitigation strategies often rely on computationally intensive language modeling tasks or rigid distance penalties, these approaches are either highly inefficient or too restrictive, heavily penalizing any deviation from the original embedding and hindering meaningful structural adaptation. To address this, the LMM objective acts as a batch-wise contrastive regularizer. Rather than forcing the adapted vector to match the original, it anchors the adapted representations to their corresponding frozen PLM representations through an InfoNCE objective \cite{oord2018representation} formulated as
\begin{equation}
\mathcal{L}_{\mathrm{LMM}} = -\frac{1}{N} \sum_{k=1}^{N} \log \frac{\exp\!\left(\frac{\tilde{\mathbf{z}}^{(k)} \cdot \mathbf{z}^{\text{frozen}(k)}}{\tau}\right)}{\sum_{j=1}^{N} \exp\!\left(\frac{\tilde{\mathbf{z}}^{(k)} \cdot \mathbf{z}^{\text{frozen}(j)}}{\tau}\right)},
\end{equation}
where $N$ denotes the number of text representations in the minibatch, $\tilde{\mathbf{z}}^{(k)}$ represents the adapted EOS representation for the $k$-th text attribute processed through an MLP projection task head, $\tau$ is set to 0.07 following standard practice in contrastive learning, and $\mathbf{z}^{\text{frozen}(k)}$ denotes the corresponding frozen PLM EOS representation before adaptation. By regularizing the latent vector with respect to the batch, this formulation ensures that an adapted representation maintains its distinct semantic identity relative to other texts. It grants the text embeddings the freedom to shift and organize around network topology, provided they remain more similar to their original semantic anchor than to the anchors of unrelated interactions. This provides a flexible and efficient mechanism to preserve linguistic structure without limiting the structural expressivity gained through DRA.

\subsection{Implementation}

We provide the hyperparameter settings of UTGA in Appendix \ref{app:hyperparameters} and text embedding extraction details, including storage and GPU compute hour requirements, in Appendix~\ref{app:embeddings}. During pretraining, UTGA jointly optimizes $\mathcal{L}_{\mathrm{DRA}}$ and $\mathcal{L}_{\mathrm{LMM}}$ using dynamic weight averaging \cite{liu2019dwa} to balance task weights based on their historical loss drop rates over recent training epochs. Performance on the validation split is monitored with a composite score:
\begin{equation}
\text{Score} = \frac{\text{MRR}_{\text{val}} - \text{MRR}_{\text{best}}}{\text{MRR}_{\text{best}}} + \frac{\text{Loss}_{\text{best}} - \text{Loss}_{\text{val}}}{\text{Loss}_{\text{best}}}.
\end{equation}
Early stopping is implemented with a patience of 10 epochs based on the composite score.

\begin{table*}[t]
\centering
\caption{Finetuning results for dynamic edge classification on DTGB datasets. Values show average Macro F1 scores in percent with subscripts denoting $\pm$ standard deviation. \textbf{Bold} indicates best performance per dataset. Relative performance gains of UTGA over the base framework DyGnROLE are shown in green superscript. *, $\dagger$, and $\ddagger$ respectively indicate statistically significant improvement at $p < 0.1$, $p < 0.05$, and $p < 0.01$ over the best-in-column baseline for individual datasets using a paired two-tailed $t$-test, and over the best overall baseline for the Avg column using a Wilcoxon signed-rank test.}
\label{tab:macro_f1_results}
\small
\setlength{\tabcolsep}{4pt}
\begin{tabular}{lccccccccc}
\toprule
\textbf{Method} & Amazon & Google & Yelp & Enron & Stack U & Stack E & GDELT & ICEWS & \textbf{Avg} \\
\midrule
Random & 15.27$_{\pm 0.04}$ & 15.03$_{\pm 0.05}$ & 16.95$_{\pm 0.03}$ & 7.01$_{\pm 0.04}$ & 44.76$_{\pm 0.05}$ & 46.72$_{\pm 0.07}$ & 0.16$_{\pm 0.01}$ & 0.13$_{\pm 0.00}$ & 18.25 \\
Majority & 15.76$_{\pm 0.00}$ & 15.93$_{\pm 0.00}$ & 13.99$_{\pm 0.00}$ & 4.06$_{\pm 0.00}$ & 44.65$_{\pm 0.00}$ & 42.82$_{\pm 0.00}$ & 0.07$_{\pm 0.00}$ & 0.12$_{\pm 0.00}$ & 17.17 \\
DyRep & 16.09$_{\pm 0.61}$ & 13.95$_{\pm 3.58}$ & 15.12$_{\pm 1.10}$ & 8.75$_{\pm 0.52}$ & 44.91$_{\pm 0.32}$ & 43.92$_{\pm 2.14}$ & 0.13$_{\pm 0.02}$ & 0.14$_{\pm 0.02}$ & 17.88 \\
JODIE & 17.25$_{\pm 1.90}$ & 17.22$_{\pm 1.35}$ & 15.17$_{\pm 2.34}$ & 6.81$_{\pm 1.48}$ & 44.90$_{\pm 0.83}$ & 42.98$_{\pm 1.63}$ & 0.11$_{\pm 0.02}$ & 0.15$_{\pm 0.02}$ & 18.07 \\
DVGMAE & 16.10$_{\pm 0.45}$ & 16.23$_{\pm 0.46}$ & 18.35$_{\pm 5.34}$ & 10.18$_{\pm 0.78}$ & 45.59$_{\pm 1.88}$ & 42.82$_{\pm 0.01}$ & 0.15$_{\pm 0.02}$ & 0.22$_{\pm 0.14}$ & 18.71 \\
IDOL & 15.91$_{\pm 0.30}$ & 15.93$_{\pm 0.00}$ & 24.80$_{\pm 0.10}$ & 8.14$_{\pm 0.25}$ & 44.65$_{\pm 0.00}$ & 42.82$_{\pm 0.00}$ & 0.12$_{\pm 0.01}$ & 0.50$_{\pm 0.18}$ & 19.11 \\
CLDG & 15.57$_{\pm 0.38}$ & 16.20$_{\pm 0.53}$ & 26.90$_{\pm 1.77}$ & 8.28$_{\pm 1.12}$ & 44.65$_{\pm 0.00}$ & 45.15$_{\pm 2.48}$ & 0.17$_{\pm 0.00}$ & 0.56$_{\pm 0.07}$ & 19.68 \\
DyGMAE & 16.05$_{\pm 0.58}$ & 15.93$_{\pm 0.00}$ & 24.86$_{\pm 0.26}$ & 8.59$_{\pm 1.10}$ & 47.12$_{\pm 2.08}$ & 48.08$_{\pm 0.41}$ & 0.19$_{\pm 0.00}$ & 0.50$_{\pm 0.07}$ & 20.17 \\
CAWN & 18.59$_{\pm 0.69}$ & 19.64$_{\pm 0.54}$ & 28.77$_{\pm 1.02}$ & 9.88$_{\pm 0.84}$ & 44.65$_{\pm 0.00}$ & 42.82$_{\pm 0.00}$ & 0.33$_{\pm 0.05}$ & 0.94$_{\pm 0.08}$ & 20.70 \\
DyGFormer & 15.76$_{\pm 0.00}$ & 18.00$_{\pm 1.67}$ & 25.69$_{\pm 2.68}$ & 9.29$_{\pm 0.42}$ & 49.68$_{\pm 4.12}$ & 50.73$_{\pm 3.16}$ & 0.36$_{\pm 0.02}$ & 1.03$_{\pm 0.03}$ & 21.32 \\
TCL & 18.52$_{\pm 0.40}$ & 19.49$_{\pm 0.44}$ & 27.85$_{\pm 0.43}$ & 9.86$_{\pm 0.38}$ & 46.35$_{\pm 3.40}$ & 48.77$_{\pm 4.93}$ & 0.38$_{\pm 0.03}$ & 0.96$_{\pm 0.04}$ & 21.52 \\
DyGnROLE & 16.53$_{\pm 0.30}$ & 19.85$_{\pm 0.54}$ & 29.35$_{\pm 0.92}$ & 10.69$_{\pm 0.44}$ & 50.99$_{\pm 2.08}$ & 51.49$_{\pm 2.16}$ & 0.38$_{\pm 0.06}$ & 1.17$_{\pm 0.09}$ & 22.56 \\
GraphMixer & 16.80$_{\pm 1.28}$ & 19.06$_{\pm 1.58}$ & 25.94$_{\pm 1.20}$ & 9.36$_{\pm 0.43}$ & 54.99$_{\pm 5.20}$ & 54.66$_{\pm 1.81}$ & 0.31$_{\pm 0.06}$ & 0.70$_{\pm 0.10}$ & 22.73 \\
TGAT & 17.99$_{\pm 1.17}$ & 19.65$_{\pm 0.87}$ & 28.97$_{\pm 0.43}$ & 9.43$_{\pm 0.28}$ & 54.82$_{\pm 0.68}$ & 55.53$_{\pm 1.10}$ & 0.36$_{\pm 0.06}$ & 1.05$_{\pm 0.05}$ & 23.48 \\
\midrule
UTGA & \textbf{24.79}$^{{\color{green!60!black}+50.0\%}\ddagger}_{\pm 0.56}$ & \textbf{25.13}$^{{\color{green!60!black}+26.6\%}\ddagger}_{\pm 0.32}$ & \textbf{36.81}$^{{\color{green!60!black}+25.4\%}\ddagger}_{\pm 0.23}$ & \textbf{18.82}$^{{\color{green!60!black}+76.1\%}\ddagger}_{\pm 1.13}$ & \textbf{55.50}$^{{\color{green!60!black}+8.8\%}}_{\pm 0.19}$ & \textbf{56.33}$^{{\color{green!60!black}+9.4\%}}_{\pm 0.75}$ & \textbf{0.63}$^{{\color{green!60!black}+65.8\%}\ddagger}_{\pm 0.03}$ & \textbf{2.26}$^{{\color{green!60!black}+93.2\%}\ddagger}_{\pm 0.17}$ & \textbf{27.53}$^{{\color{green!60!black}+22.0\%}\ddagger}$ \\
\bottomrule
\end{tabular}
\end{table*}

\section{Experiments}
\label{sec:experiments}

In this section, we investigate the impact of semantic and topological co-evolution on dynamic edge classification. We outline our experimental setup and evaluate our framework alongside diverse baselines on multiple datasets.

\subsection{Setup}
\label{sec:setup}

\paragraph{Datasets.} We evaluate the unified text-graph framework on all eight datasets from the Dynamic Text-Attributed Graph Benchmark (DTGB) \cite{zhang2024dtgb}. These datasets feature timestamped edges accompanied by raw text attributes, spanning four domains: e-commerce reviews (Amazon, Google, Yelp), technical Q\&A forums (Stack E, Stack U), corporate communications (Enron), and temporal knowledge graphs (GDELT, ICEWS). The datasets are available through the official DTGB repository\footnote{\url{https://github.com/zjs123/DTGB}} as chronologically ordered edge lists with source and destination node IDs, timestamps, and edge labels. Detailed descriptions, summary statistics, label distributions, and average node and edge text lengths are provided in Appendix \ref{app:dataset_details}.

Crucially, because our investigation centers on unified text-graph alignment, DTGB provides a rigorous setting due to the substantial variation in the length of its node and edge text attributes. Average text length varies considerably across datasets and between modalities. Node attributes are particularly long in the technical Q\&A datasets, averaging 142 words for Stack E and 109 words for Stack U, while they are much shorter in Enron and GDELT, averaging only 1 and 2 words, respectively. The extremely short node attributes in Enron arise because they consist of email addresses rather than natural-language descriptions. Edge attributes exhibit a different pattern: Enron, Yelp, and Amazon contain comparatively long interaction text, averaging 151, 105, and 96 words, respectively, whereas the temporal knowledge graphs contain very short edge descriptions, averaging only 4 words for both GDELT and ICEWS. Google similarly provides relatively short text in both modalities, with averages of 8 words for nodes and 23 words for edges. These differences expose the framework to substantially different distributions of linguistic information, from text-rich node attributes and long-form interactions to very short textual descriptions, allowing us to evaluate whether structural information can be effectively integrated across diverse semantic conditions.

\paragraph{Tasks.}Performance is assessed across four downstream tasks corresponding to the dataset domains. On the e-commerce networks (Amazon, Google, Yelp), models perform \textbf{rating prediction} to predict a 1-5 star rating accompanying a future review. On the technical Q\&A forums (Stack E, Stack U), the task is \textbf{post usefulness prediction}, determining if a future reply will be upvoted by the community. For corporate communications (Enron), models perform \textbf{email topic classification}, categorizing future messages into one of 10 subjects (e.g., ``calendar'', ``deal''). On the temporal knowledge graphs (GDELT, ICEWS), the models are evaluated on \textbf{political relation classification}, a highly complex task requiring the model to predict future geopolitical interactions from a large label space (236 categories for GDELT, 266 for ICEWS).

\paragraph{Baselines.}To provide a comprehensive comparison, we benchmark our framework against fourteen models. These baselines span fundamental heuristic predictors (Random, Majority) and established CTDGNs, including memory-centric models \cite{kumar2019jodie, trivedi2019dyrep}, those using attention mechanisms or random walks \cite{xu2020tgat, wang2021cawn, wang2021tcl, yu2023dygformer}, and an MLP-based architecture \cite{cong2023graphmixer}. Furthermore, we include recent self-supervised approaches designed for dynamic graphs, covering both generative masked autoencoders \cite{liu2025dygmae, gao2025dvgmae} and contrastive learning methods \cite{xu2023cldg, zhu2024idol, bonnet2026dygnrole}. All baseline models use text representations computed by Qwen3-Embedding-0.6B \cite{zhang2025qwen3} to ensure fair comparison with UTGA. Detailed hyperparameter configurations are provided in Appendix \ref{app:hyperparameters}.

\subsection{Training Procedure}
\label{sec:training_procedure}

We follow the standard 70-15-15 chronological training-validation-test split defined by DTGB. All methods with a self-supervised pretraining phase first pretrain on the 200,000 most recent interactions of the training split and monitor performance on the 30,000 most recent interactions of the validation set. Consistent with UTGA, early stopping of self-supervised baseline models uses a patience of 10 epochs during pretraining.

Self-supervised models are then finetuned while supervised models are trained from random initialization for dynamic edge classification. Following the standard DTGB evaluation protocol, the node representations generated by the models are concatenated and passed through an MLP classifier consisting of two hidden layers with ReLU activations and dropout regularization, followed by a linear output layer producing class logits. The full architecture and classification head are trained end-to-end using cross-entropy loss. We perform this experiment with varying label budgets for dynamic edge classification (1k, 5k, 10k, and 20k training interactions, with proportional 150, 750, 1,500, and 3,000 validation interactions, respectively), holding out the full DTGB test split for final evaluation. Early stopping uses a 20-epoch patience and an initial 20-epoch grace period based on improvements in the validation Macro F1 score.

\subsection{Dynamic Edge Classification Performance}
\label{sec:future_edge_classification}

Table \ref{tab:macro_f1_results} reports dynamic edge classification results at the 10,000-label budget. UTGA achieves the highest Macro F1 on every dataset and on average (27.53\%), a 22.0\% relative gain over the base framework DyGnROLE (22.56\%) and 17.2\% over the strongest overall baseline, TGAT (23.48\%). Since DyGnROLE isolates the effect of text-graph co-evolution, we use it as the primary reference in the discussion below, and we name the best-in-column baseline wherever it is a different model. The improvement over the best-in-column baseline is statistically significant ($p < 0.01$) on six of the eight datasets.

On Amazon, Google, and Yelp, UTGA improves over DyGnROLE by 50.0\%, 26.6\%, and 25.4\%, respectively. DyGnROLE is also the best-in-column baseline on Google and Yelp. On Amazon the best baseline is CAWN (18.59\%), over which UTGA achieves a gain of 33.4\%. On Enron, UTGA scores 18.82\%, a 76.1\% relative improvement over DyGnROLE (10.69\%), which is also the best-in-column baseline.

Stack U and Stack E expose the vulnerability of standard architectures to class imbalance, with the majority class comprising roughly 81\% and 75\% of interactions, respectively. Several advanced models fail to meaningfully surpass the majority heuristic (44.65\% and 42.82\% Macro F1). On these datasets, UTGA reaches 55.50\% and 56.33\%, improving over DyGnROLE by 8.8\% and 9.4\%. Its margin over the best-in-column baselines (GraphMixer on Stack U, TGAT on Stack E) is narrower (+0.9\% and +1.4\%). The per-class F1 on the minority ``useful'' class separates the strongest baselines and UTGA more clearly (Table \ref{tab:stack_class1_results}). Here the framework outperforms the best-in-column baselines by 13.2\% (Stack U) and 6.3\% (Stack E), and DyGnROLE by 94.9\% and 54.4\%, while exhibiting tighter standard deviations across random seeds.

\begin{table}[t]
\centering
\caption{Minority class (``useful'') F1 scores on the Stack Exchange datasets. Values show average F1 scores in percent with subscripts denoting $\pm$ standard deviation. \textbf{Bold} indicates best performance per dataset. Relative performance gains of UTGA over the base framework DyGnROLE are shown in green superscript.}
\label{tab:stack_class1_results}
\small
\setlength{\tabcolsep}{6pt}
\begin{tabular}{lccc}
\toprule
\textbf{Method} & Stack U & Stack E & \textbf{Avg} \\
\midrule
DyGnROLE & 15.16$_{\pm 5.46}$ & 21.75$_{\pm 6.22}$ & 18.45 \\
GraphMixer & 26.11$_{\pm 13.22}$ & 31.12$_{\pm 7.48}$ & 28.61 \\
TGAT & 24.45$_{\pm 1.98}$ & 31.60$_{\pm 4.20}$ & 28.02 \\
\midrule
UTGA & \textbf{29.55}$^{{\color{green!60!black}+94.9\%}}_{\pm 0.90}$ & \textbf{33.59}$^{{\color{green!60!black}+54.4\%}}_{\pm 3.25}$ & \textbf{31.57}$^{{\color{green!60!black}+71.1\%}}$ \\
\bottomrule
\end{tabular}
\end{table}

\begin{figure}[t]
    \centering
    \includegraphics[width=\columnwidth]{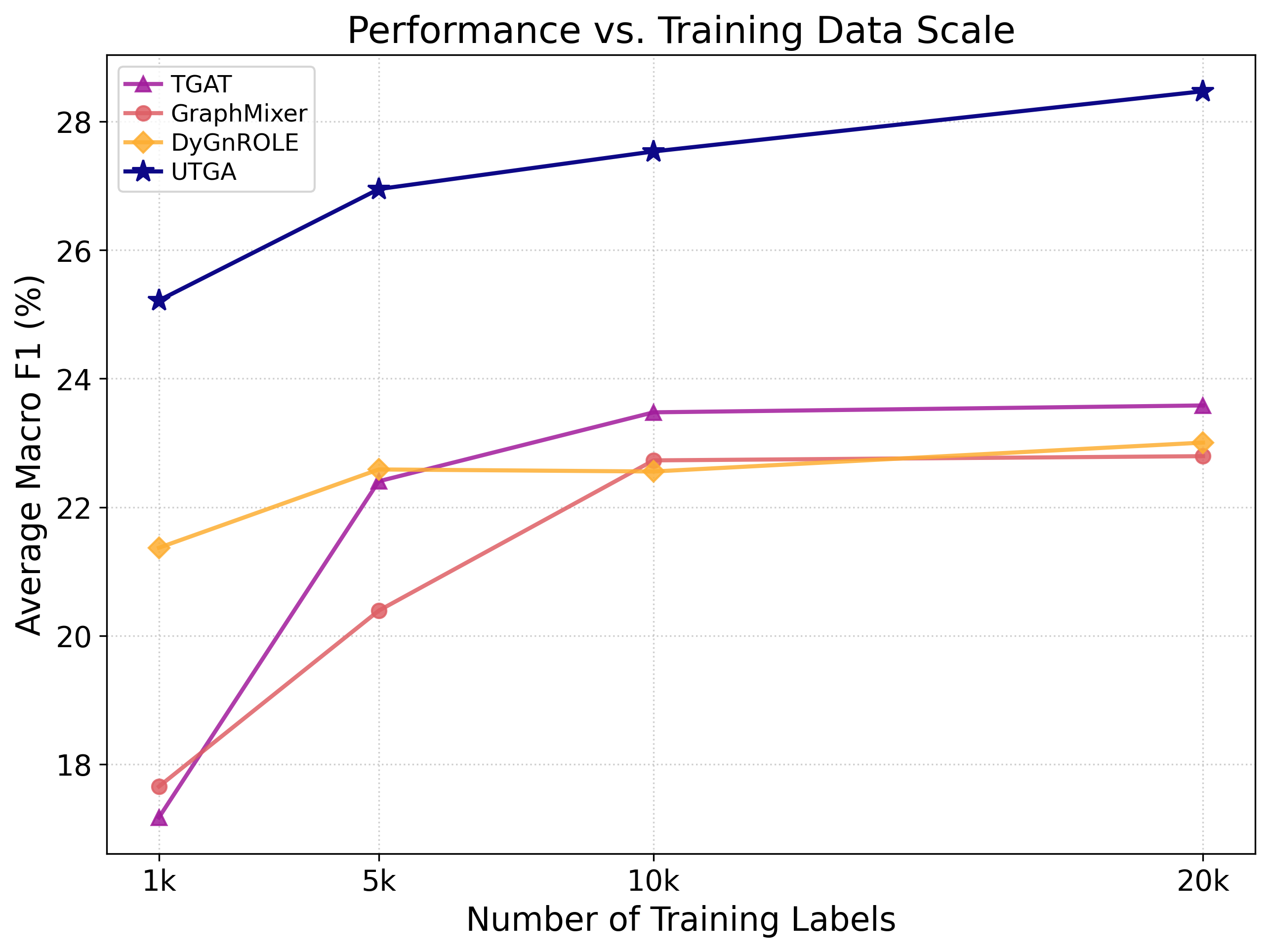}
    \caption{Average Macro F1 performance across the DTGB benchmark as the number of training labels scales from 1k to 20k. UTGA maintains a consistent and substantial advantage over the top three performing baselines (TGAT, GraphMixer, and DyGnROLE) at every data scale.}
    \Description{A line graph plotting Average Macro F1 percentage against the number of training labels at 1k, 5k, 10k, and 20k intervals for UTGA and the top 3 baseline models.}
    \label{fig:performance_scale}
\end{figure}

GDELT and ICEWS are the most challenging settings, with massive label vocabularies and very short text attributes. On GDELT, UTGA reaches 0.63\%, a 65.8\% improvement over DyGnROLE, which is tied with TCL as the best-in-column baseline (0.38\%). On ICEWS, UTGA achieves 2.26\%, a 93.2\% relative gain over DyGnROLE (1.17\%), again the best-in-column baseline, nearly doubling its performance.

Overall, these observations indicate that allowing the textual latent space to co-evolve with the graph topology during pretraining yields a superior and more robust foundation for downstream classification across a highly diverse set of text-attributed domains.

\begin{table*}[t]
\centering
\small
\caption{Ablation results for dynamic edge classification showing average Macro F1 scores (\%). \textbf{Bold} indicates best on dataset.}
\label{tab:ablation}
\setlength{\tabcolsep}{4pt}
\begin{tabular}{lccccccccc}
\toprule
\textbf{UTGA Variants} & Amazon & Google & Yelp & Enron & Stack U & Stack E & GDELT & ICEWS & \textbf{Avg} \\
\midrule
w/o Pretraining & 18.57 $\pm$ 0.42 & 15.93 $\pm$ 0.00 & 30.24 $\pm$ 1.35 & 8.89 $\pm$ 0.71 & 47.64 $\pm$ 3.68 & 47.60 $\pm$ 5.90 & 0.32 $\pm$ 0.05 & 1.77 $\pm$ 0.08 & 21.37 \\
w/o LMM & 23.40 $\pm$ 1.02 & 24.06 $\pm$ 0.35 & 31.35 $\pm$ 0.44 & 9.08 $\pm$ 0.77 & 54.66 $\pm$ 1.06 & \textbf{56.53 $\pm$ 0.60} & 0.56 $\pm$ 0.01 & 2.07 $\pm$ 0.12 & 25.21 \\
MLP PLM Adapter & 21.91 $\pm$ 0.25 & 21.45 $\pm$ 0.40 & 34.17 $\pm$ 0.35 & \textbf{19.23 $\pm$ 1.13} & 53.52 $\pm$ 1.05 & 55.64 $\pm$ 1.02 & 0.43 $\pm$ 0.04 & 1.24 $\pm$ 0.06 & 25.95 \\
\midrule
UTGA & \textbf{24.79 $\pm$ 0.56} & \textbf{25.13 $\pm$ 0.32} & \textbf{36.81 $\pm$ 0.23} & 18.82 $\pm$ 1.13 & \textbf{55.50 $\pm$ 0.19} & 56.33 $\pm$ 0.75 & \textbf{0.63 $\pm$ 0.03} & \textbf{2.26 $\pm$ 0.17} & \textbf{27.53} \\
\bottomrule
\end{tabular}
\end{table*}

\begin{figure}[t]
    \centering
    \includegraphics[width=\columnwidth]{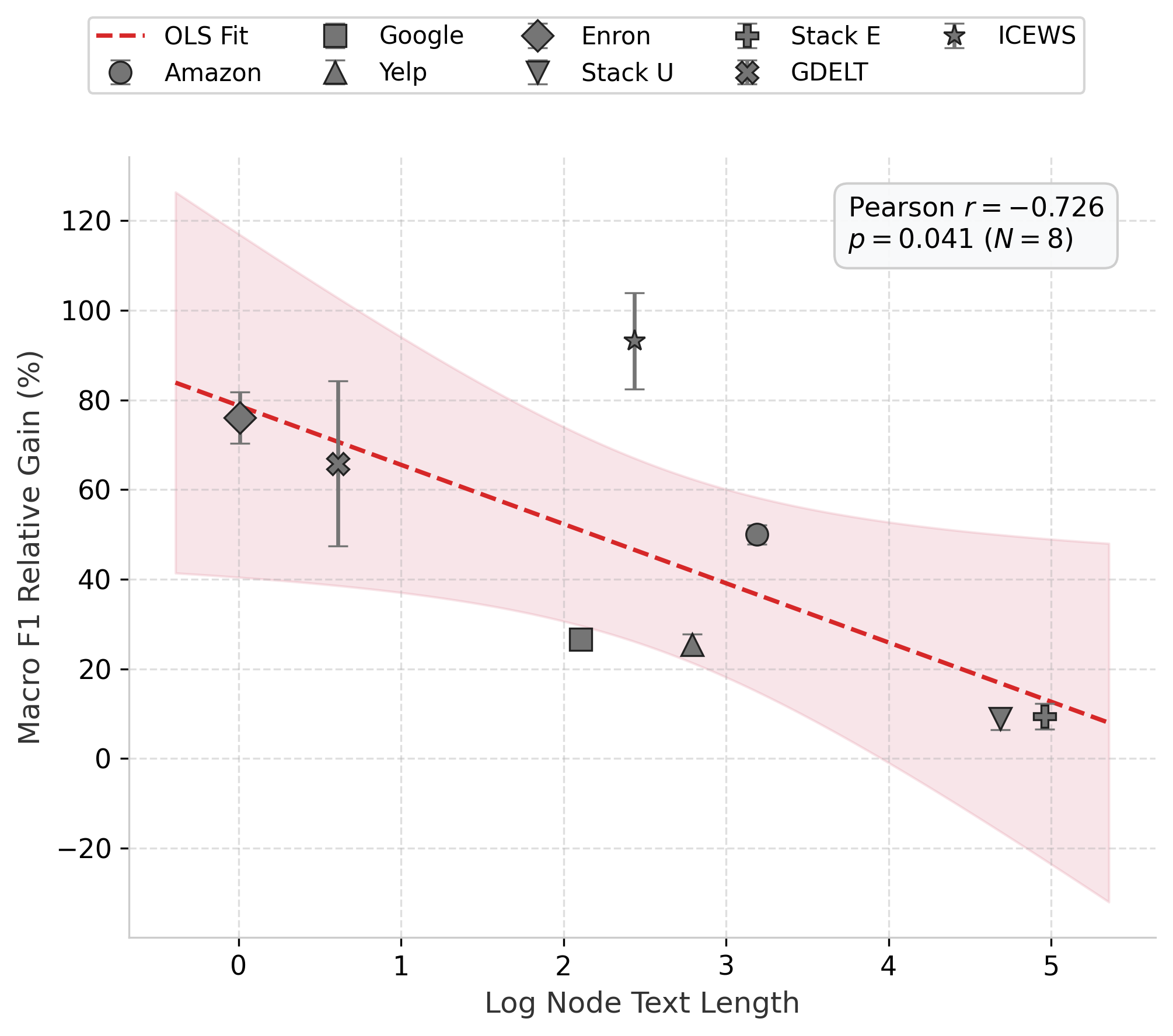}
    \caption{Relationship between log-transformed average node text length and relative Macro F1 performance gain of UTGA over the base framework. The red dashed line represents the Ordinary Least Squares (OLS) fit with a 95\% confidence interval band. Markers display dataset-level means with standard error bars.}
    \Description{A scatter plot showing a negative correlation between log node text length on the x-axis and Macro F1 relative gain percentage on the y-axis.}
    \label{fig:text_length_vs_gain}
\end{figure}

\textbf{Robustness to label availability.} To evaluate the robustness of the learned representations under varying degrees of label availability, we scale the finetuning data budget from 1,000 to 20,000 training interactions. Figure \ref{fig:performance_scale} plots the average Macro F1 score across all datasets for UTGA and the top three performing baselines: TGAT, GraphMixer, and DyGnROLE. UTGA demonstrates superior sample efficiency, maintaining a dominant performance margin at every scale. Notably, UTGA trained with only 1,000 labels (25.1\% Macro F1) outperforms all three baseline models when they are trained with 20,000 labels.

\textbf{Impact of text length.} As visualized in Figure \ref{fig:text_length_vs_gain}, revisiting the text length variations discussed in Section \ref{sec:setup} reveals an empirical pattern regarding where text-graph co-evolution provides the most value. We observe a strong, statistically significant negative correlation between log-transformed average node text length and relative Macro F1 gain at the aggregated dataset level ($N=8$; Pearson $r = -0.726$, $p = 0.041$). This analysis suggests that semantic and spatio-temporal co-evolution yields the highest relative performance improvements in environments with short node text.

Because datasets with shorter node texts frequently feature larger label spaces in this benchmark (dataset details in Appendix \ref{app:dataset_details}), we conducted a confounding analysis to ensure this performance gain is not merely an artifact of task difficulty. To separate the effect of text length from the high class counts (200+) of GDELT and ICEWS, we performed a subgroup analysis restricted to the other six datasets, which have 10 or fewer classes. In this controlled setting, the negative relationship intensifies and remains statistically significant at the aggregated dataset level ($N=6$; Pearson $r = -0.869$, $p = 0.024$).

\section{Analysis}
\label{sec:analysis}

In this section, we analyze the specific mechanisms driving the impact of semantic and topological co-evolution. This includes isolating component contributions, measuring semantic anisotropy, and probing the structural expressivity of the text embeddings.

\subsection{Ablation Study}
\label{sec:ablation}

To isolate the effects of pretraining objectives and the PLM adapter of UTGA, we conduct an ablation study. We evaluate three variants of the architecture: (1) \textbf{w/o Pretraining}, where the DRA and LMM objectives are removed, the model is initialized with random weights for dynamic edge classification, and the PLM Adapter operates on the token-level Qwen3 embeddings as in the complete model; (2) \textbf{w/o LMM}, where the LMM objective is dropped during pretraining, allowing the full architecture to update solely via structural DRA gradients; and (3) \textbf{MLP PLM Adapter}, where the Transformer-based PLM Adapter is replaced by an MLP-based PLM Adapter, which still adapts the semantic space, but it operates purely on the pooled EOS representations of the texts with an MLP rather than adapting the full token sequence with a Transformer.

\begin{figure}[t]
    \centering
    \includegraphics[width=\columnwidth]{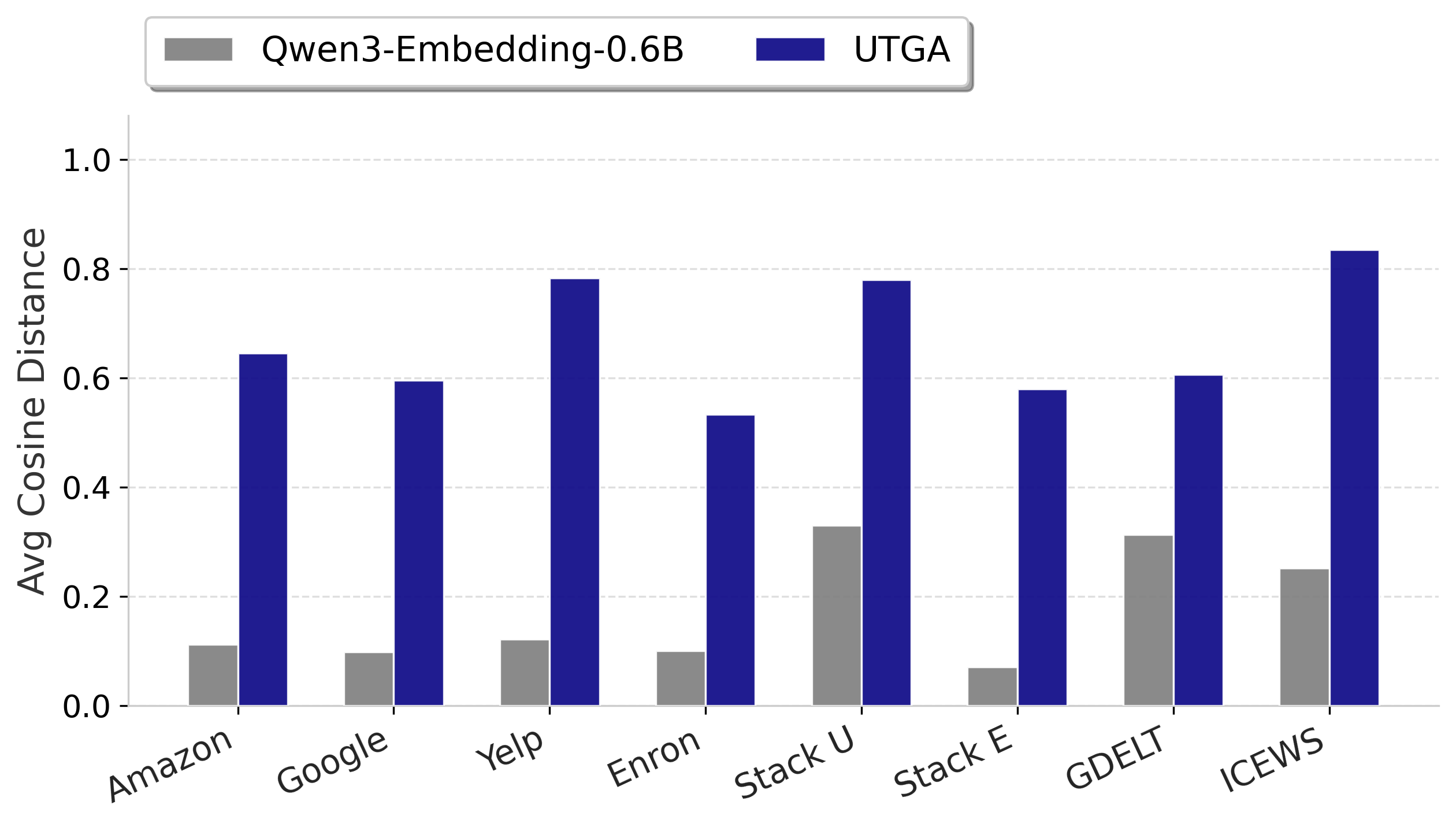}
    \caption{Average pairwise cosine distance for edge text embeddings. The representations of the raw Qwen3-Embedding-0.6B exhibit high levels of anisotropy, whereas the UTGA framework effectively reduces this collapse to expand the embedding space.}
    \Description{A grouped bar chart comparing the average cosine distance of edge text embeddings before and after applying the UTGA framework across the DTGB datasets.}
    \label{fig:embedding_spread}
\end{figure}

The results in Table \ref{tab:ablation} highlight the necessity of the joint pretraining framework. Removing pretraining entirely causes the average Macro F1 to fall from 27.53\% to 21.37\%, confirming the benefits of initializing with a PLM Adapter that encodes structural and temporal properties into the text embeddings. Removing the LMM objective results in a drop to 25.21\%. Without this anchor, the structural gradients from DRA optimize solely for topology, which limits the semantic utility of the representations and reduces downstream classification performance. Substituting the Transformer for an MLP in the PLM Adapter drops overall performance to 25.95\%, demonstrating that adapting the full sequence of text tokens generally provides a better alignment than adapting a single pooled sentence representation. While the complete UTGA framework scores marginally lower than MLP PLM Adapter on Enron and w/o LMM on Stack E, these specific variations fall within the standard deviations of the complete model ($\pm$ 1.13 and $\pm$ 0.75, respectively), indicating no evidence of a difference on these datasets rather than a systematic disadvantage.

\begin{table*}[t]
\centering
\small
\caption{Node structural probing analysis on GDELT comparing the linear separability (Macro F1 \%) of node text embeddings on graph structural properties. \textbf{Bold} indicates best performance per task.}
\label{tab:node_probing_gdelt}
\setlength{\tabcolsep}{4pt}
\begin{tabular*}{\textwidth}{@{\extracolsep{\fill}}lccccc}
\toprule
\textbf{Embedding} & \textbf{In-Degree} & \textbf{Out-Degree} & \textbf{PageRank} & \textbf{Clustering} & \textbf{Lifespan} \\
\midrule
Random & 25.09 $\pm$ 0.58 & 24.09 $\pm$ 0.46 & 25.31 $\pm$ 0.65 & 33.04 $\pm$ 0.82 & 21.29 $\pm$ 0.35 \\
Qwen3-Embedding & 23.75 $\pm$ 0.08 & 23.93 $\pm$ 3.53 & 27.69 $\pm$ 0.35 & 32.90 $\pm$ 4.83 & 22.43 $\pm$ 0.00 \\
\midrule
UTGA & \textbf{30.02 $\pm$ 3.91} & \textbf{29.55 $\pm$ 1.88} & \textbf{38.85 $\pm$ 0.68} & \textbf{44.86 $\pm$ 6.05} & \textbf{27.03 $\pm$ 1.54} \\
\bottomrule
\end{tabular*}
\end{table*}

\subsection{Anisotropy Reduction}
\label{sec:embedding_spread}

General-purpose language models frequently map diverse texts into a narrow, anisotropic cone, causing even semantically unrelated texts to exhibit artificially low cosine distances \cite{ethayarajh2019how, li2020sentence}. This geometric homogeneity makes it difficult for downstream graph encoders to differentiate distinct interactions. To evaluate how our framework mitigates this degradation, we compute the average off-diagonal pairwise cosine distance for uniformly sampled subsets of up to 10,000 test-split edges using Qwen3-Embedding-0.6B and the UTGA framework (pretrained, but not finetuned). As shown in Figure \ref{fig:embedding_spread}, raw Qwen3 embeddings exhibit severe anisotropy, with average distances frequently around 0.1. The UTGA framework effectively mitigates this by increasing cosine distances substantially across all datasets.

To confirm the anisotropy reduction provided by UTGA yields more discriminable features, we probe the linear separability of the frozen edge representations of the e-commerce datasets by predicting 1-5 star review ratings using the same training and validation data sampled for pretraining and evaluating on the full test splits. The results in Table \ref{tab:edge_probing} show that the narrow geometry of the raw Qwen3 embeddings restricts the spatial resolution required to separate sentiments, limiting performance. By expanding the embedding space while preserving semantic integrity via LMM, UTGA constructs representations that are both highly discriminable and linguistically meaningful, achieving substantially higher linear separability across all three datasets.

\begin{table}[t]
\centering
\small
\caption{Edge probing analysis comparing the linear separability (Macro F1 \%) of edge text embeddings on e-commerce datasets. \textbf{Bold} indicates best performance per dataset.}
\label{tab:edge_probing}
\setlength{\tabcolsep}{4pt}
\begin{tabular}{lccc}
\toprule
\textbf{Embedding} & Amazon & Google & Yelp \\
\midrule
Random & 15.27 $\pm$ 0.04 & 15.03 $\pm$ 0.05 & 16.95 $\pm$ 0.03 \\
Qwen3-Embedding & 18.99 $\pm$ 0.13 & 23.79 $\pm$ 0.05 & 24.22 $\pm$ 0.23 \\
\midrule
UTGA & \textbf{36.71 $\pm$ 0.52} & \textbf{35.69 $\pm$ 0.53} & \textbf{44.28 $\pm$ 0.56} \\
\bottomrule
\end{tabular}
\end{table}

\subsection{Topological Alignment}
\label{sec:topological_alignment}

A central hypothesis of our investigation is that the UTGA framework encodes network topology into the text embeddings. We evaluate the structural expressivity of the learned representations through a series of node text embedding probing tasks. Using the GDELT temporal knowledge graph as a case study, we extract five fundamental graph-theoretic properties for each node: in-degree, out-degree, PageRank centrality \cite{page1999pagerank}, temporal lifespan, and local clustering density. The first four are divided into four quartile tiers while local clustering density is categorized into three classes: zero local triangles, below the non-zero median, and at or above the non-zero median. The discretization thresholds are calculated on the 200,000-interaction pretraining split. We train a single linear layer classifier on the frozen node text embeddings using the same training and validation data sampled for pretraining. Evaluation is performed on the full test splits.

The results of the probing analysis are presented in Table \ref{tab:node_probing_gdelt}. As expected, the raw Qwen3 features lack structural awareness, yielding classification scores near or below random guessing on almost every task. When the representations are updated through UTGA, performance improves on all tasks. The framework yields substantially higher linear separability on PageRank centrality and local clustering density, indicating that the adapted representations encode both global network positioning and local topological motifs.

\subsection{Computational Efficiency}
\label{sec:efficiency}

To quantify the overhead of our framework, we compare UTGA to the base framework (DyGnROLE) and a joint text-graph training baseline that replaces the PLM Adapter with Low-Rank Adaptation (LoRA) \cite{hu2022lora} on the full Qwen3-Embedding-0.6B backbone. As reported in Table~\ref{tab:efficiency}, joint training with LoRA restricts throughput to 28.53~samples/s (4,485.74~ms per batch) and raises peak GPU memory allocation to 4,335.87~MB.

In contrast, with 2.03M additional trainable parameters over DyGnROLE (3.58M total), UTGA reaches a throughput of 1,600.62 samples/s (79.76~ms per batch). This yields a 56$\times$ speedup and an 88.8\% memory reduction (485.34~MB) compared to LoRA. While slower than static feature processing (21.24~ms per batch), UTGA enables dynamic semantic-structural co-evolution within the same memory footprint order of magnitude. Details regarding our computing hardware are provided in Appendix~\ref{app:compute}.

\begin{table}[t]
\centering
\small
\caption{Computational efficiency on the Enron dataset. Metrics include trainable parameters (Params), average per-batch execution time in milliseconds (Time), overall throughput (Samples/s), and peak GPU memory allocation in megabytes (Mem) during pretraining.}
\label{tab:efficiency}
\resizebox{\columnwidth}{!}{%
\begin{tabular}{lcccc}
\toprule
\textbf{Architecture} & \textbf{Params} & \textbf{Time} & \textbf{Samples/s} & \textbf{Mem} \\
\midrule
DyGnROLE & 1.55M & 21.24 $\pm$ 6.21 & 5980.29 & 171.45 \\
UTGA w/ LoRA & 6.86M & 4485.74 $\pm$ 1743.91 & 28.53 & 4335.87 \\
\midrule
UTGA & 3.58M & 79.76 $\pm$ 14.61 & 1600.62 & 485.34 \\
\bottomrule
\end{tabular}%
}
\end{table}

\section{Conclusion}
\label{sec:conclusion}

In this work, we propose a unified text-graph alignment (UTGA) framework to investigate the effects of enabling the co-evolution of text representations and network topology in dynamic text-attributed graphs. With this framework, we demonstrate that updating token-level text representations via spatio-temporal gradients consistently improves dynamic edge classification performance over standard static-feature baselines, particularly while simultaneously anchoring them with a language model maintenance objective to prevent detrimental semantic distortion. Through representational analyses, we revealed that UTGA reduces anisotropy, improving the linear separability of the embeddings, and effectively encodes structural and temporal properties into the semantic space. Crucially, a computational efficiency analysis confirms that text-graph co-evolution via UTGA is achieved without incurring the prohibitive computational overhead typical of end-to-end joint training.

\bibliographystyle{ACM-Reference-Format}
\bibliography{references}

\appendix

\section{Dataset Details}
\label{app:dataset_details}

\begin{table}[h]
    \centering
    \caption{Summary statistics of the DTGB datasets.}
    \label{tab:dtgb_stats_table}
    \begin{tabular}{l@{\hskip 0.2in}r@{\hskip 0.2in}r@{\hskip 0.2in}r@{\hskip 0.2in}r}
    \toprule
    Dataset & \#Nodes & \#Edges & \#Classes & \#Words \\
    \midrule
    Amazon & 293,566 & 3,217,324 & 5 & 623,447,701 \\
    Enron & 42,711 & 797,907 & 10 & 187,576,324 \\
    GDELT & 6,786 & 1,339,245 & 236 & 7,632,528 \\
    Google & 111,168 & 1,380,623 & 5 & 74,401,703 \\
    ICEWS & 31,796 & 1,100,071 & 266 & 27,003,734 \\
    Stack E & 397,702 & 1,262,225 & 2 & 340,401,237 \\
    Stack U & 674,248 & 1,497,006 & 2 & 347,878,722 \\
    Yelp & 2,138,242 & 6,990,189 & 5 & 987,141,388 \\
    \bottomrule
    \end{tabular}
\end{table}

\begin{figure}[t]
    \centering
    \includegraphics[width=\columnwidth]{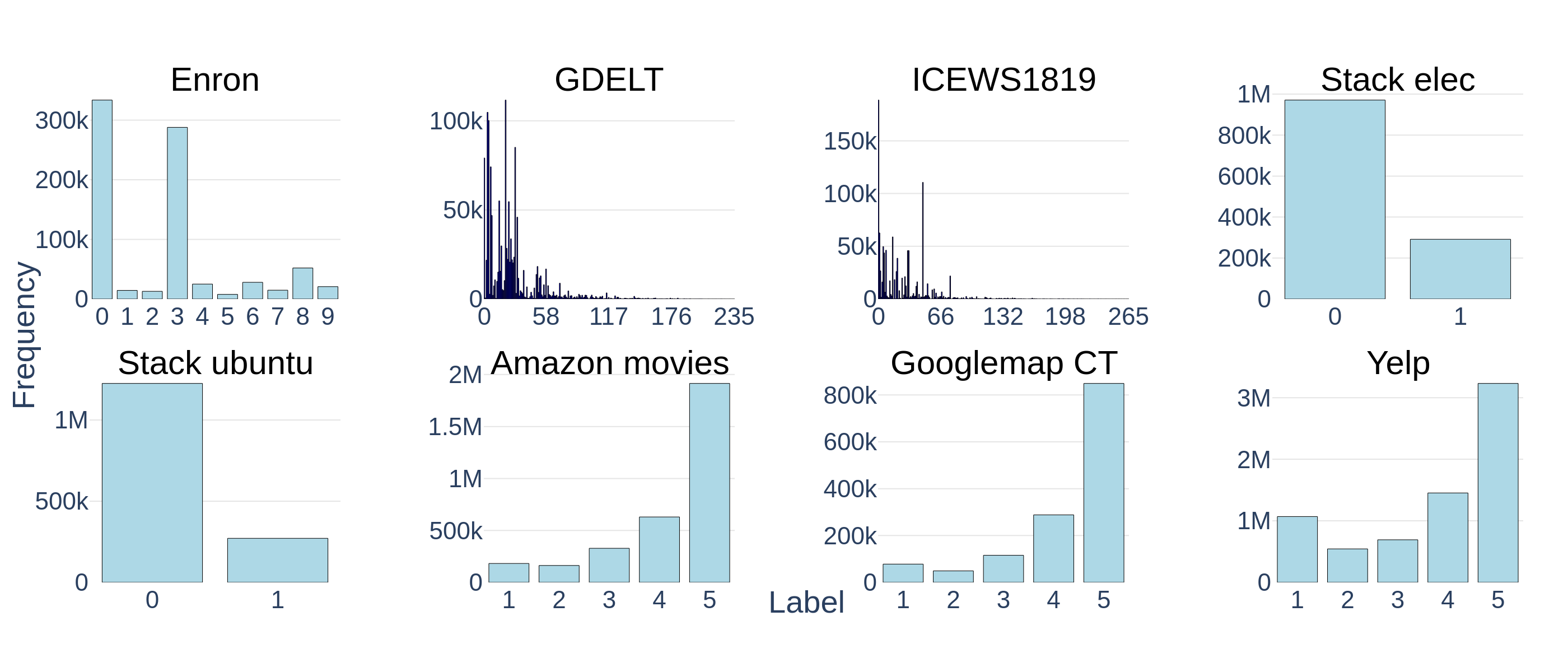}
    \caption{Label distributions of the DTGB datasets.}
    \Description{Eight bar charts showing the frequency of different classes for the DTGB datasets.}
    \label{fig:dtgb_label_distributions}
\end{figure}

\begin{figure}[t]
    \centering
    \includegraphics[width=\columnwidth]{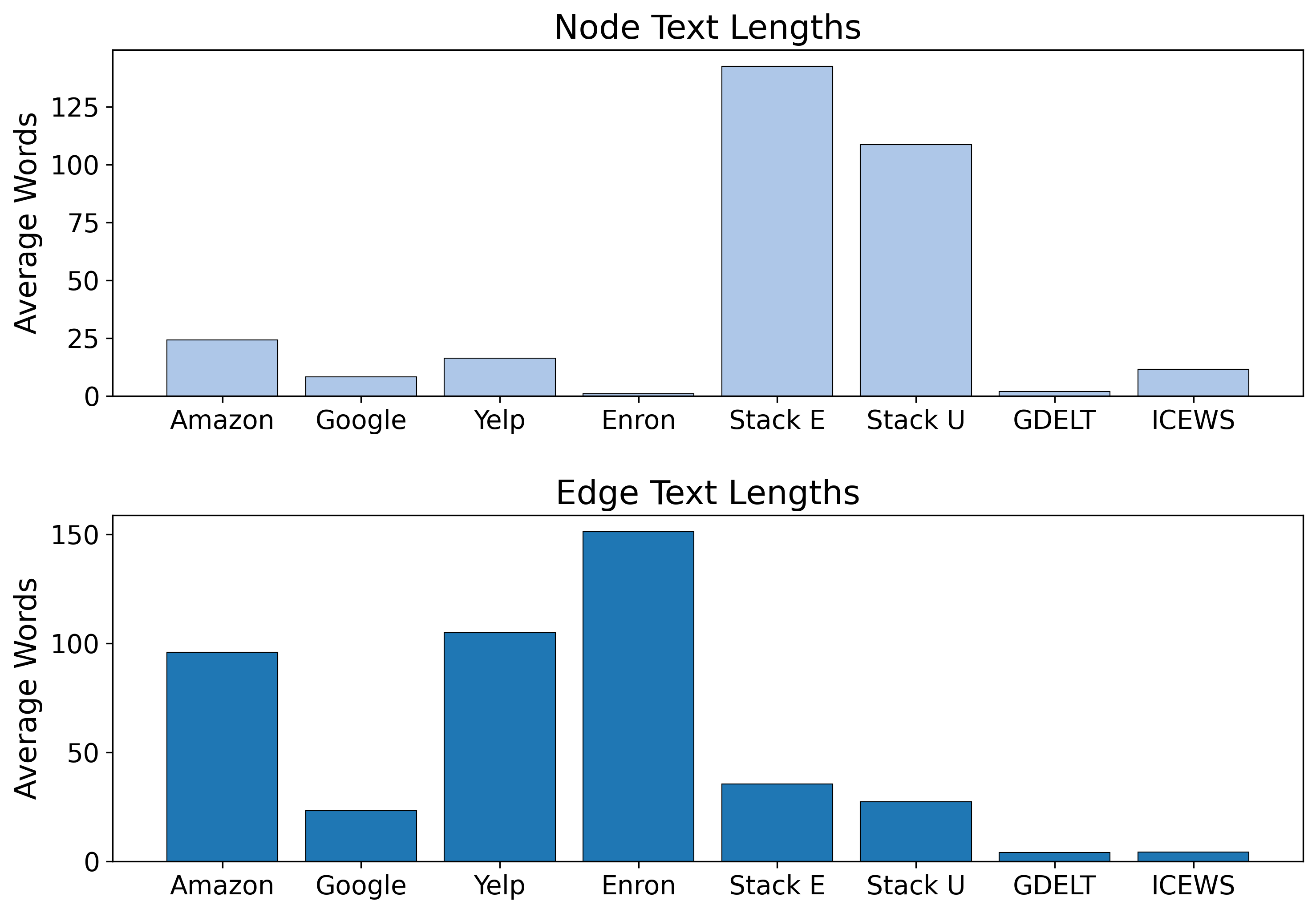}
    \caption{Average node and edge text lengths of the DTGB datasets.}
    \Description{A two-panel bar chart displaying the average word count for node and edge texts across the DTGB datasets.}
    \label{fig:dtgb_text_lengths}
\end{figure}

\textbf{Enron}\footnote{\url{https://www.cs.cmu.edu/~enron/}}: Emails between Enron employees from 1999 to 2002.
\begin{itemize}
    \item \textbf{Nodes}: Employees
    \item \textbf{Node texts}: Professional email addresses
    \item \textbf{Edges}: Email exchanges between employees
    \item \textbf{Edge texts}: Raw email content
    \item \textbf{Edge labels}: 10 distinct communication categories (e.g., 'deal', 'calendar')
\end{itemize}

\textbf{GDELT}\footnote{\url{https://www.gdeltproject.org/}}: Global Database of Events, Language, and Tone knowledge graph representing geopolitical events.
\begin{itemize}
    \item \textbf{Nodes}: Political entities or actors
    \item \textbf{Node texts}: Entity names
    \item \textbf{Edges}: Relationships or actions between two political entities
    \item \textbf{Edge texts}: Concise descriptions of the relationship
    \item \textbf{Edge labels}: 236 unique relationship types (e.g., 'provide economic aid')
\end{itemize}

\textbf{ICEWS1819 (ICEWS)}\footnote{\url{https://dataverse.harvard.edu/dataverse/icews}}: Integrated Crisis Early Warning System knowledge graph representing geopolitical events.
\begin{itemize}
    \item \textbf{Nodes}: Political entities
    \item \textbf{Node texts}: Entity names
    \item \textbf{Edges}: Documented relationships between political actors
    \item \textbf{Edge texts}: Short descriptions of the event or relationship
    \item \textbf{Edge labels}: 266 unique interaction categories (e.g., 'Engage in negotiation')
\end{itemize}

\textbf{Stack elec (Stack E)}\footnote{\url{https://archive.org/details/stackexchange}}: Stack Exchange forum interactions focused on electronics and electrical engineering.
\begin{itemize}
    \item \textbf{Nodes}: Users or specific questions
    \item \textbf{Node texts}: User introduction and location: question title and post content
    \item \textbf{Edges}: The submission of an answer or a comment on a post
    \item \textbf{Edge texts}: Textual content of answers or comments
    \item \textbf{Edge labels}: Binary classification ('useful' or 'useless') based on a voting score where $score = upvotes - downvotes$. An edge is 'useful' if $score > 1$.
\end{itemize}

\textbf{Stack ubuntu (Stack U)}\footnote{\url{https://archive.org/details/stackexchange}}: Stack Exchange forum interactions focused on the Ubuntu operating system.
\begin{itemize}
    \item \textbf{Nodes}: Users or questions
    \item \textbf{Node texts}: User introduction and location: question title and post content
    \item \textbf{Edges}: Answer submissions or comments on existing posts
    \item \textbf{Edge texts}: Textual content of answers or comments
    \item \textbf{Edge labels}: Binary classification ('useful' or 'useless') where $score = upvotes - downvotes$ and 'useful' requires $score > 1$.
\end{itemize}

\textbf{Googlemap CT (Google)}\footnote{\url{https://mcauleylab.ucsd.edu/public_datasets/gdrive/googlelocal/}}: Reviews for businesses in the state of Connecticut on Google.
\begin{itemize}
    \item \textbf{Nodes}: Users or local businesses
    \item \textbf{Node texts}: Usernames for individuals: business name, address, category, and introduction
    \item \textbf{Edges}: Review interactions between users and businesses
    \item \textbf{Edge texts}: The textual content of user reviews
    \item \textbf{Edge labels}: 1 to 5 based on the standard five-star rating system
\end{itemize}

\textbf{Amazon movies (Amazon)}\footnote{\url{https://cseweb.ucsd.edu/~jmcauley/datasets/amazon_v2/}}: Reviews for the Movies and TV category on Amazon.
\begin{itemize}
    \item \textbf{Nodes}: Users or products in the Movies and TV category
    \item \textbf{Node texts}: Usernames for individuals: product title, description, category, and rank score
    \item \textbf{Edges}: Reviews submitted by users for specific products
    \item \textbf{Edge texts}: The textual content of user reviews
    \item \textbf{Edge labels}: 1 to 5 based on the five-star rating system
\end{itemize}

\textbf{Yelp}\footnote{\url{https://www.yelp.com/dataset}}: Reviews for businesses on Yelp.
\begin{itemize}
    \item \textbf{Nodes}: Users or businesses
    \item \textbf{Node texts}: Name, register date, review count, and average rating for users: business name, city, address, and category
    \item \textbf{Edges}: Review interactions between users and businesses
    \item \textbf{Edge texts}: The textual content of user reviews
    \item \textbf{Edge labels}: 1 to 5 based on the five-star rating system
\end{itemize}

\section{Model Hyperparameters}
\label{app:hyperparameters}

\textbf{Shared Settings}
\begin{itemize}
    \item Optimizer: AdamW
    \item Maximum number of neighbors ($S$): 10
    \item Gradient clipping max norm: 1.0
    \item Max training epochs: 1000
    \item Pretraining random seed: 42
    \item Pretraining early stopping patience: 10 epochs
    \item Finetuning random seeds: 0, 1, 2, 3, 4
    \item Finetuning early stopping patience: 20 epochs
    \item PyTorch automatic mixed precision FP16
\end{itemize}

\textbf{UTGA}
\begin{itemize}
    \item Weight decay: 0.01
    \item Batch size: 128
    \item Pretraining learning rate: $5\times 10^{-5}$
    \item Finetuning learning rate: $3\times 10^{-5}$
    \item Token feature dimension: 256
    \item Maximum tokens per text: 256
    \item PLM Adapter layers: 2
    \item PLM Adapter heads: 4
    \item PLM Adapter dropout rate: 0.1
    \item PLM Adapter SwiGLU hidden dimension: 768
    \item PLM Adapter RMSNorm $\epsilon$: $10^{-6}$
    \item Transformer encoder layers: 2
    \item Transformer encoder heads: 2
    \item Transformer encoder dropout rate: 0.1
    \item Transformer encoder feed-forward dimension: 1000
    \item Channel dimension: 50
    \item Time dimension: 100
    \item Number of channels: 5
    \item Maximum number of neighbors ($S$): 10
    \item LMM projection head: $256 \rightarrow 512 \rightarrow 256$, dropout 0.1
    \item DWA temperature: 2.0
    \item DWA warm-up epochs: 2
\end{itemize}

\textbf{DyGnROLE}
\begin{itemize}
    \item Transformer encoder layers: 2
    \item Transformer encoder heads: 2
    \item Transformer encoder dropout rate: 0.1
    \item Channel dimension: 50
    \item Time dimension: 100
    \item Occurrence embedding dimension: 50
    \item Number of channels: 5
    \item InfoNCE temperature ($\tau$): 0.07
\end{itemize}

\textbf{CLDG}
\begin{itemize}
    \item InfoNCE temperature: 0.07
    \item Hidden dimension: 128
    \item Embedding dimension: 64
    \item Encoder layers: 2
    \item Dropout rate: 0.1
    \item Number of snapshots: 5
\end{itemize}

\textbf{IDOL}
\begin{itemize}
    \item Hidden dimension: 512
    \item Embedding dimension: 128
    \item Neighborhood samples ($K$): 4
    \item Restart probability ($\alpha$): 0.1
    \item Maximum bias ($r_{max}^b$): 0.01
    \item Loss trade-off ($\lambda$): 0.01
    \item Validation temperature: 0.07
\end{itemize}

\textbf{DVGMAE}
\begin{itemize}
    \item Hidden dimension: 128
    \item Embedding dimension: 32
    \item Dropout rate: 0.1
    \item Epsilon ($\epsilon$): 0.25
    \item Mask ratio: 0.5
    \item Loss lambda weight ($\lambda$): 0.5
    \item Number of snapshots: 5
\end{itemize}

\textbf{DyGMAE}
\begin{itemize}
    \item Hidden dimension: 128
    \item Encoder layers: 2
    \item Dropout rate: 0.1
    \item Number of masks: 2
    \item Mask ratios: 0.2, 0.4
    \item Mask types: path, random
    \item Fusion type: attention
    \item Walk length: 2
    \item Loss gamma ($\gamma$): 0.1
    \item Loss delta ($\delta$): 0.1
    \item Loss lambda $c$ ($\lambda_c$): 0.1
    \item Number of snapshots: 5
\end{itemize}

\textbf{TGAT}
\begin{itemize}
    \item Layers: 2
    \item Heads: 2
    \item Dropout rate: 0.1
    \item Time dimension: 100
    \item Maximum number of neighbors ($S$): 10
\end{itemize}

\textbf{DyRep}
\begin{itemize}
    \item Memory dimension: 128
    \item Mailbox size: 10
    \item Feature projection: $256 \rightarrow 128$
    \item Memory reset: each epoch
\end{itemize}

\textbf{JODIE}
\begin{itemize}
    \item Memory dimension: 128
    \item Recurrent unit: single-layer RNN cell
    \item Time dimension: 128
    \item Dropout rate: 0.1
    \item Memory reset: each epoch
\end{itemize}

\textbf{GraphMixer}
\begin{itemize}
    \item Mixer layers: 2
    \item Number of tokens: 10
    \item Channel dimension: 256
    \item Time dimension: 100 (fixed encoder)
    \item Token expansion factor: 0.5
    \item Channel expansion factor: 4.0
    \item Dropout rate: 0.5
\end{itemize}

\textbf{TCL}
\begin{itemize}
    \item Transformer encoder layers: 2
    \item Transformer encoder heads: 2
    \item Transformer encoder dropout rate: 0.1
    \item Hidden dimension: 256
    \item Feed-forward dimension: 1024
    \item Time dimension: 100
    \item Number of depth embeddings: 11
    \item Maximum number of neighbors ($S$): 10
\end{itemize}

\textbf{CAWN}
\begin{itemize}
    \item Walk length: 2
    \item Neighbors per hop: 10
    \item Attention heads: 8
    \item Dropout rate: 0.1
    \item Time dimension: 100
    \item Position encoding dimension: 172
    \item Feature BiLSTM hidden dimension: 392
    \item Position BiLSTM hidden dimension: 86
\end{itemize}

\textbf{DyGFormer}
\begin{itemize}
    \item Transformer encoder layers: 2
    \item Transformer encoder heads: 2
    \item Transformer encoder dropout rate: 0.1
    \item Transformer encoder feed-forward dimension: 800
    \item Channel dimension: 50
    \item Time dimension: 100
    \item Neighbor co-occurrence dimension: 50
    \item Maximum sequence length ($S$): 10
\end{itemize}

\section{Text Embedding Extraction}
\label{app:embeddings}

The text attributes are embedded at the token level using Qwen3-Embedding-0.6B \cite{zhang2025qwen3} and truncated to 256 dimensions to balance representational capacity with computational and memory efficiency. Across all datasets, the active token cache comprises 607.2M tokens (289.6 GiB in FP16 at D = 256), with text sequences capped at a maximum of 256 tokens. The embeddings are stored in a ragged CSR binary format, discarding all padding tokens. Feature extraction required ~10 GPU-hours. Details regarding our computing hardware are provided in Appendix~\ref{app:compute}.

\section{Computing Hardware}
\label{app:compute}

All experiments were performed using an Intel Xeon Silver 4214 CPU (2.20GHz), 128 GB of system RAM, and four NVIDIA GeForce RTX 2080 Ti GPUs (11 GB VRAM each).

\section{Limitations and Future Work}
\label{app:limit_future}

While our investigation establishes the empirical benefits of anchored text-graph co-evolution for a diverse set of dynamic edge classification tasks, extending this framework to link prediction and node classification remains a valuable direction for future research. A compelling question regarding link prediction is whether a dedicated pretraining phase remains advantageous, given that the structural supervision signal (i.e., the existence of edges) is naturally abundant, unlike the often limited annotations typical of edge and node classification. Consequently, subsequent studies could contrast the efficacy of the two-stage pretraining paradigm against an end-to-end framework where the PLM Adapter and graph encoder are jointly optimized from random initialization using only the downstream objective, as we have done with our ``w/o Pretraining'' ablation.

\end{document}